\pdfoutput=1

\documentclass[11pt]{article}

\usepackage[preprint]{coling}

\usepackage{times}
\usepackage{latexsym}

\usepackage[T1]{fontenc}

\usepackage[utf8]{inputenc}

\usepackage{microtype}

\usepackage{inconsolata}

\usepackage{graphicx}

\usepackage[utf8]{inputenc}
\usepackage{textgreek}
\usepackage{amsmath} 
\usepackage{tabularray}

\title{HLB: Benchmarking LLMs’ Humanlikeness in Language Use}

\author{
\textbf{Xufeng Duan\textsuperscript{1}}\quad
\textbf{Bei Xiao\textsuperscript{1}}\quad
\textbf{Xuemei Tang\textsuperscript{1}}\quad
\textbf{Zhenguang G. Cai\textsuperscript{1,2}}\\
\textsuperscript{1}Department of Linguistics and Modern Languages, The Chinese University of Hong Kong\\
\textsuperscript{2}Brain and Mind Institute, The Chinese University of Hong Kong\\
\texttt{xufeng.duan@link.cuhk.edu.hk}
}

\begin{document}
\maketitle
\begin{abstract}
As synthetic data becomes increasingly prevalent in training language models, particularly through generated dialogue, concerns have emerged that these models may deviate from authentic human language patterns, potentially losing the richness and creativity inherent in human communication. This highlights the critical need to assess the humanlikeness of language models in real-world language use. In this paper, we present a comprehensive humanlikeness benchmark (HLB) evaluating 20 large language models (LLMs) using 10 psycholinguistic experiments designed to probe core linguistic aspects, including sound, word, syntax, semantics, and discourse (see \href{https://huggingface.co/spaces/XX}{this link}). To anchor these comparisons, we collected responses from over 2,000 human participants and compared them to outputs from the LLMs in these experiments.

For rigorous evaluation, we developed a coding algorithm that accurately identified language use patterns, enabling the extraction of response distributions for each task. By comparing the response distributions between human participants and LLMs, we quantified humanlikeness through distributional similarity. Our results reveal fine-grained differences in how well LLMs replicate human responses across various linguistic levels. Importantly, we found that improvements in other performance metrics did not necessarily lead to greater humanlikeness, and in some cases, even resulted in a decline. By introducing psycholinguistic methods to model evaluation, this benchmark offers the first framework for systematically assessing the humanlikeness of LLMs in language use.
\end{abstract}

\section{Introduction}

In recent years, large language models (LLMs) have made significant advancements. Models like OpenAI’s GPT series and Meta’s Llama family can generate human-like text, engage in coherent dialogues, and answer complex questions, often producing responses that are indistinguishable from those of humans in certain evaluations \cite{tsubota_text_2024}. \citet{cai_large_2024} conducted a systematic evaluation of human-like language use in models such as ChatGPT and Vicuna, demonstrating that LLMs closely replicate human language patterns in many aspects. However, despite these successes, questions remain about how accurately these models capture the deeper, nuanced patterns of human language use. In other words, the full extent of their similarity to human behavior remains unclear.

The importance of evaluating humanlikeness in language use is further underscored by the increasing reliance on synthetic data for model training, particularly in dialogue models. While synthetic data generation facilitates efficient scaling of model training, it raises concerns about models diverging from real-world human language patterns\cite{10.1093/pnasnexus/pgae400}. Studies have shown that synthetic data can degrade model performance after retraining\cite{shumailov2024ai}. This makes it imperative to assess the humanlikeness of LLMs rigorously across various aspects of language use, to ensure that models do not lose the diversity and richness of human language data.

To address this challenge, we introduce a psycholinguistic benchmark designed to provide a systematic and comprehensive evaluation of how closely LLMs align with human linguistic behavior.

Although numerous benchmarks and leaderboards have been developed to assess the performance of LLMs on downstream NLP tasks, they often fail to capture the finer, human-like qualities of language use. Current NLP benchmarks typically focus on task-based accuracy or performance \cite{lewkowycz_solving_2022,zhou_solving_2023,peng_humaneval-xl_2024,hendrycks_measuring_2021,zellers_hellaswag_2019}, overlooking the broader psycholinguistic dimensions that characterize how humans process and produce language. Furthermore, few studies have systematically compared the language use of LLMs and human participants across multiple linguistic levels. This gap highlights the need for a new benchmark that can robustly measure the extent to which LLMs replicate human language behavior in real-world, diverse linguistic contexts.

In this paper, we address this gap by presenting a psycholinguistic benchmark study that evaluates the humanlikeness of 20 LLMs. Our benchmark consists of 10 representative psycholinguistic experiments, adapted from \citet{cai_large_2024}, which cover five core linguistic aspects: sound, word, syntax, semantics, and discourse, with two experiments dedicated to each aspect (see \ref{tab:experiments}). We collected approximately 50 to 100 responses per item from over 2,000 human participants. Additionally, we gathered 100 responses per item from each of the 20 LLMs, including well-known models such as GPT-4o, GPT-3.5, Llama 2, Llama 3, Llama 3.1, and other state-of-the-art models (see Table~\ref{tab:experiments}). To quantify humanlikeness, we developed an auto-coding algorithm that efficiently and reliably extracts language use patterns from responses. The humanlikeness metric was then calculated based on the similarity between the response distributions of humans and LLMs, using a comparison of their probability distributions.

Our findings reveal significant, nuanced differences in how LLMs perform across various linguistic aspects, offering a new benchmark for evaluating the humanlikeness of LLMs in natural language use. This benchmark introduces psycholinguistic methods to model evaluation and provides the first framework for systematically assessing the humanlikeness of LLMs in language use.
\begin{figure}[t]
      \centering
      \includegraphics[width=\linewidth]{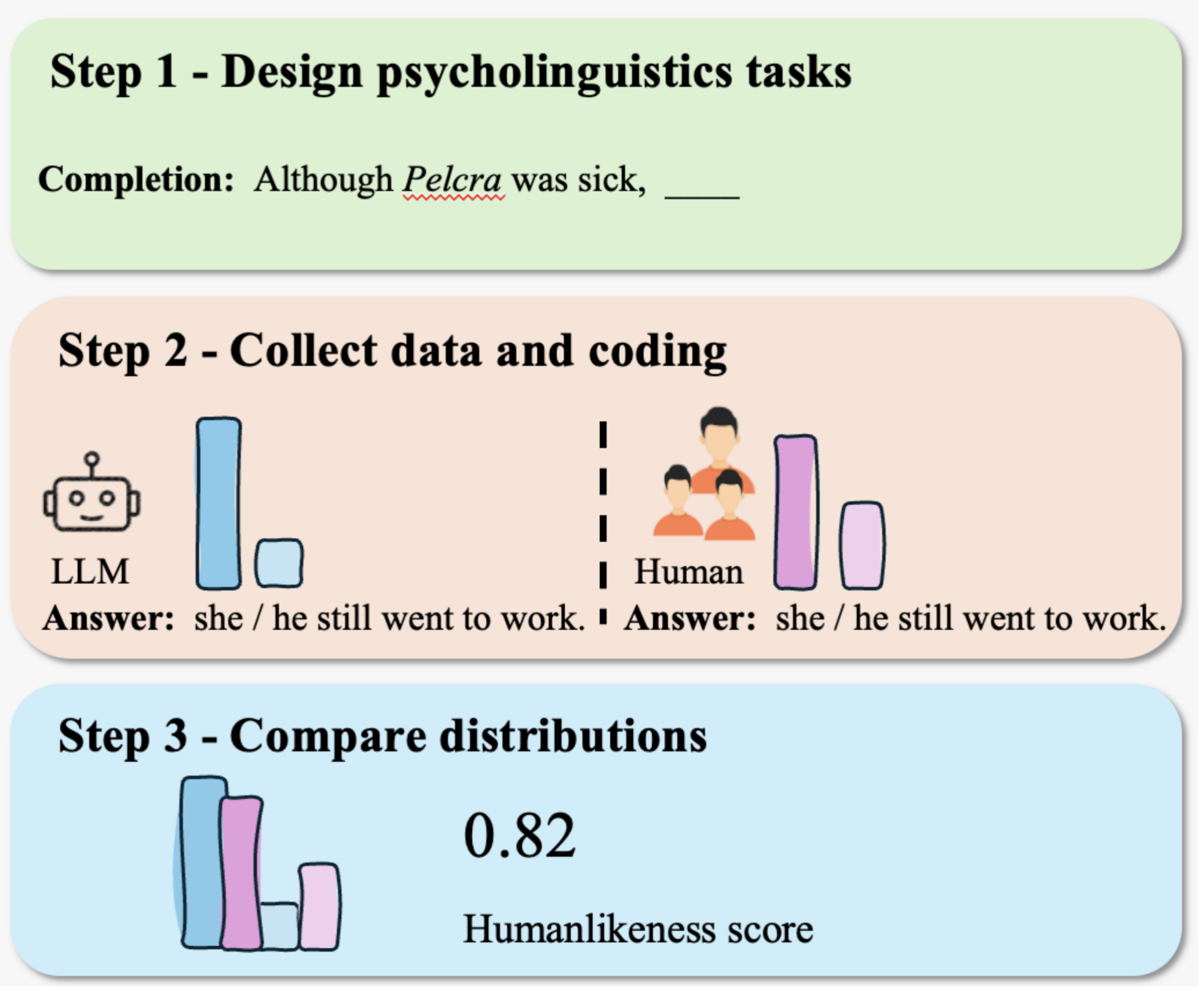}
      \caption{The benchmark framework. The example prompt is taken from the sound-gender association task, where humans can infer the gender of a novel name (e.g., \textit{Pelcra} for Female; \textit{Pelcrad} for Male) based on phonology.}
      \label{fig:enter-label}
  \end{figure}

\begin{table}[h]
\centering
\begin{tblr}{
  cell{2}{2} = {r=2}{},
  cell{4}{2} = {r=2}{},
  cell{6}{2} = {r=2}{},
  cell{8}{2} = {r=2}{},
  cell{10}{2} = {r=2}{},
  hline{1-2,12} = {-}{},
  hline{4,6,8,10} = {2-3}{},
  colspec = {p{0.35cm} p{1.2cm} p{4.4cm}},
}
Exp & Level     & Task                                \\
E1      & Sound     & sound-shape association \cite{kohler1967gestalt}             \\
E2      &           & sound-gender association \cite{cassidy1999inferring}         \\
E3      & Word      & word length and predictivity \cite{mahowald2013info}        \\
E4      &           & word meaning priming \cite{rodd2013long}               \\
E5      & Syntax    & structural priming \cite{pickering1998representation}                 \\
E6      &           & syntactic ambiguity resolution \cite{altmann1988interaction}     \\
E7      & Meaning   & implausible sentence interpretation \cite{gibson2013rational} \\
E8      &           & semantic illusion \cite{erickson1981words}                   \\
E9      & Discourse & implicit causality \cite{garvey1974implicit}                  \\
E10     &           & drawing inferences \cite{singer2015phantom}                  
\end{tblr}
\caption{The experiments in this benchmark.}
\label{tab:experiments}
\end{table}

\section{Related Work}
Recent advances in LLMs have led to the development of various benchmarks designed to evaluate their linguistic capabilities. Standard benchmarks like GLUE \cite{wang2018glue} and SuperGLUE \cite{wang2019superglue} assess models across a range of natural language processing (NLP) tasks, including sentence classification, textual entailment, and question answering. However, these benchmarks primarily focus on task-based accuracy and often overlook the more intricate aspects of humanlike language processing. While these evaluations provide valuable insights into model performance, they do not fully capture the extent to which LLMs comprehend and generate language in a humanlike manner. As \citet{manning_emergent_2020} note, LLMs are powerful statistical models that can identify patterns in vast datasets, but these benchmarks do not adequately test how well models replicate human patterns of language use due to the interplay of complex cognitive biases.
\subsection{Psychological Experimentation on LLMs}
A growing body of research has begun applying classical psychological experiments to evaluate LLMs in more domain-specific and cognitively demanding tasks. For example, \citet{binz_using_2023} and \citet{dasgupta_language_2023} used well-known psychological paradigms, such as the Linda problem and the Wason selection task, to probe LLMs’ abilities in judgment and decision-making. Similarly, \citet{sap_neural_2023} and \citet{trott_large_2023} explored whether LLMs exhibit theory of mind, a key component of human social cognition, while \citet{miotto_who_2022} and \citet{karra_estimating_2023} examined LLMs’ personality traits. In the domain of behavioral economics, Horton (2023) conducted experiments with GPT-3 to explore its decision-making processes.
These studies suggest that LLMs can be treated as cognitive agents in psychological experiments, providing insights into how LLMs align with humans in reasoning, behavior, and decision-making. Moreover, they help shed light on the underlying mechanisms of LLMs, as seen in the work of \citet{huang_towards_2023}and \citet{qiao_reasoning_2023}, who analyzed reasoning patterns in LLMs. \citet{hagendorff_machine_2023} further provided a comprehensive review of LLM performance in psychological tests, showing that while LLMs demonstrate sophisticated behaviors, they often diverge from human cognition. These divergences highlight the need for more robust frameworks to understand the limitations of LLMs in mimicking human thought processes.
\subsection{Psycholinguistic Experimentation on LLMs}
Psycholinguistic approaches offer a deeper analysis by testing LLMs on how well they replicate the cognitive processes underlying human language processing. \citet{ettinger2020bert} and \citet{futrell2019neural} have subjected models like BERT to psycholinguistic tasks such as syntactic ambiguity resolution and structural priming, revealing both the strengths and limitations of LLMs in replicating human language processing. \citet{michaelov_emergent_2023} used structural priming tasks to investigate how LLMs internalize syntactic structures, while \citet{huang_large-scale_2024} H examined LLMs’ ability to resolve syntactic ambiguity. \citet{qiu_pragmatic_2023} explored how well LLMs handle pragmatic reasoning.
These studies demonstrate that LLMs can, to some extent, mimic humanlike behavior in controlled experiments. However, divergences in processing reveal the distinctions between machine learning models and humans. A recent review by \citet{demszky_using_2023} emphasized the need for benchmarks that incorporate psychological paradigms to evaluate LLMs. The authors argue that by applying psycholinguistic methods, researchers can better understand how closely LLMs approximate human cognition and where they fall short.
Despite extensive research on LLMs’ performance across various tasks, there is still no benchmark that includes human language processing data to reveal the extent to which LLMs resemble humans, particularly in language use. This paper addresses that gap by adapting 10 psycholinguistic experiments from \citet{cai_large_2024} to evaluate how closely LLMs align with human language behavior, covering phenomena ranging from sound symbolism to discourse comprehension.
\section{Methodology}

\subsection{Human Experiments}

\textbf{Experimental Design} The human experiments were conducted using Qualtrics, an online survey platform \cite{qualtrics2024}. The study included ten psycholinguistic tasks that spanned various linguistic levels, from sound, word, syntax, and meaning to discourse comprehension, with two experiments for each level (see Appendix~\ref{sec:appendix} for details). We exposed a participant to only one trial on each experiment, with a total of 10 trials across all the experiments. This setup minimized trial-level effects and facilitated direct comparisons with LLMs, which were tested under similar conditions (presenting instructions and stimuli in a single prompt) to avoid context effects within individual conversations.

\textbf{Procedure} After providing consent, participants completed the ten psycholinguistic tasks (presented in a random order); four attention checks were randomly interspersed among the trials to later identify participants for random responding. Each experimental task began with an instructional screen, some of which included examples to clarify task requirements. The examples were carefully designed to differ from the experimental stimuli to prevent potential priming effects. For instance, in a sentence-completion task, an illustrative example that did not resemble the experimental stimuli and did not induce target words for any stimuli was used. The priming tasks (which included pairs of priming and target stimuli) were spread across multiple pages to avoid strategic responses in case participants realise the relation between the prime and the target. The overall experimental procedure was streamlined for clarity and efficiency, with each session lasting approximately 8 to 10 minutes (mean = 8.336, \textit{SD} = 4.171).

\textbf{Participants} Participants were recruited from the crowd-sourcing platform Quatrics and restricted to native English speakers residing in the UK and US, according to their registration on Prolific. They were required to use a desktop computer to complete the tasks. Among the  2,205 participants taking part in the experiments, 290 were excluded for not well adhering to the experimental instructions, including completing the study too quickly, showing low effort, or not finishing the experiment, according to the Qualtrics system. The remaining 1,915 participants were further checked for language nativeness and their accuracy with attention checks. After a thorough screening process—excluding those who were not native speakers, failed attention checks, or exhibited irregularities such as excessively short completion times or multiple participation attempts—the final valid sample consisted of 1,905 participants. The sample was composed of participants as follows: female (n = 1,051), male (n = 838), preferred not to disclose (n = 16), with an average age of 44.8 years (range: 18 to 89 years). Educational levels included: no formal education (n = 2), elementary school (n = 12), high school (n = 672), bachelor’s degree (n = 862), and master’s degree (n = 357). This sample of participants resulted in each item being tested in a minimum average of 24 trials (e.g., Word Length and Predictability) and up to an average of 96 trials (e.g., Sound-Shape Association Task).

\subsection{LLM Experiments}

\textbf{Experimental Design} To compare human responses with those generated by LLMs, we employed the same 10 psycholinguistic tasks designed for human participants. 20 LLMs (See Table~\ref{tab:models}) were selected for evaluation, including models from prominent families like OpenAI’s GPT series (GPT-4o, GPT-3.5), Meta’s Llama series (Llama 2, Llama 3, Llama 3.1) and Mistral series\cite{openai_gpt-4_2024,touvron_llama_2023,mistral2024}. Each model provided 100 responses per item in each experiment, ensuring that the response data was comparable to the human data. Similar to the human experimental design, LLMs followed a one-trial-per-run paradigm, ensuring that responses were generated independently for each item to prevent context effects. The input format for the LLMs closely mirrored the instructions provided to human participants. Careful modification of human prompts was performed to ensure that task instructions were clear and interpretable by LLMs. This allowed for a direct comparison between human and LLM performance on the same tasks under identical conditions.

\textbf{Response Collection Procedure} This closely mirror that in the human experiments. Each LLM was presented with the task instructions and the stimulus combined into a single prompt. We collected 100 responses (across different conditions) for each item in an experiment in order to ensure a sufficiently large dataset for robust analysis of the response distributions. For OpenAI models, responses were obtained through the OpenAI API, while models hosted on Hugging Face were accessed using the Hugging Face Inference API. All requests to the models were made using their default parameters to encourage variability in responses. The collected responses were stored and processed for subsequent coding and analysis.

\subsection{Response Coding}

\textbf{Development and Validation} We employed an auto-coding algorithm across 10 experiments to assess agreement between human annotations and machine-generated labels. This algorithm utilized spaCy’s \textit{en\_core\_web\_trf-3.7.3} model for syntactic parsing (e.g., structural priming and syntactic ambiguity resolution tasks) and regular expressions to detect answer patterns in others. Across 20,953 trials of human response data, we computed Cohen’s Kappa (\textkappa), a measure that corrects for chance agreement between the results from manually coding and auto-coding algorithm, defined as:
\begin{equation}
  \label{eq:example}
  K = \frac{P_0 - P_e}{1 - P_e}
\end{equation}

where  Po  is the observed agreement, and  Pe  is the expected agreement by chance. 

The Kappa score was κ = 0.993, indicating near-perfect agreement (\textit{z} = 451, \textit{p} < 0.001). This demonstrates the high accuracy of the auto-coding algorithm in replicating human annotations.

\subsection{Humanlikeness Scoring}
To quantify the humanlikeness of LLM responses, we used Jensen-Shannon (JS) divergence to compare the response distributions between human participants and LLMs. JS divergence, a symmetric measure of similarity between two probability distributions, is ideal for assessing how closely LLM responses mirror human behavior across linguistic levels. For each task, the auto-coding algorithm generated response distributions for both humans and LLMs. We computed \textbf{humanlikeness score (HS)} for each item as:
\begin{equation}
    \begin{aligned}
        HS_{\text{item}} &= 1 - JS(P, Q) \\
        &= 1 - \frac{1}{2} \left[ KL(P \parallel M) + KL(Q \parallel M) \right]
    \end{aligned}
\end{equation}

where P and Q are the human and LLM response distributions, and M is their average. For each experiment, we average the scores across all items. The overall humanlikeness score across all experiments is then computed as:
\begin{equation} 
    \begin{aligned}
        HS_{\text{Overall}} &= \frac{1}{m} \sum_{j=1}^{m} \left( \frac{1}{n_j} \sum_{i=1}^{n_j} \left( 1 - \frac{1}{2} \left[ \text{KL}(P_i \parallel M_i) \right.\right.\right. \\
        &\qquad\qquad\qquad \left.\left.\left. + \text{KL}(Q_i \parallel M_i) \right] \right) \right)
    \end{aligned}
\end{equation}

\begin{figure*}[t]
    \centering
    \includegraphics[width=\textwidth]{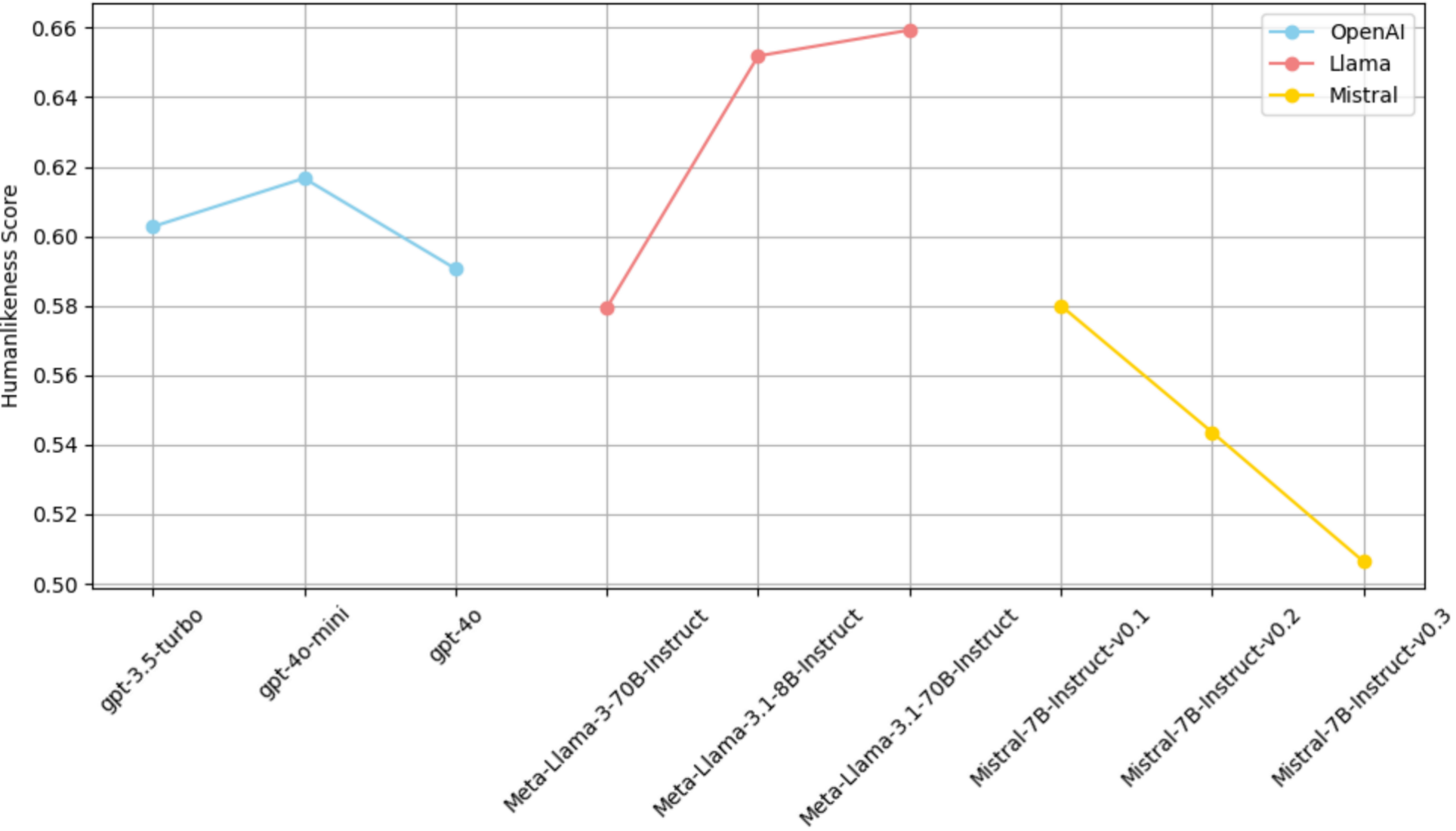}
    \caption{Humanlikeness scores of three LLM families}
    \label{fig:enter-label}
\end{figure*}

\begin{table*}[t]
\centering
\begin{tblr}{
  cell{1}{1} = {r=2}{},
  cell{1}{2} = {r=2}{},
  cell{1}{3} = {c=2}{},
  cell{1}{5} = {c=2}{},
  cell{1}{7} = {c=2}{},
  cell{1}{9} = {c=2}{},
  cell{1}{11} = {c=2}{},
  hline{1,3,24} = {-}{},
  hline{2} = {3-12}{},
}
Experiment                  & Overall & Sound &    & Word &    & Meaning &    & Syntax &    & Discourse &     \\
                            &         & E1    & E2 & E3   & E4 & E5      & E6 & E7     & E8 & E9        & E10 \\
Meta-Llama-3.1-70B-Instruct & 66.50   & 89    & 62 & 61   & 6  & 81      & 77 & 80     & 67 & 80        & 63  \\
Meta-Llama-3.1-8B-Instruct  & 65.89   & 73    & 65 & 60   & 12 & 84      & 78 & 79     & 74 & 79        & 56  \\
Phi-3-mini-4k-instruct      & 64.61   & 61    & 68 & 59   & 19 & 89      & 71 & 76     & 48 & 80        & 76  \\
Mistral-Nemo-Instruct-2407  & 63.69   & 74    & 63 & 56   & 10 & 84      & 77 & 52     & 75 & 79        & 68  \\
Llama-2-13b-chat-hf         & 63.15   & 57    & 57 & 51   & 25 & 75      & 67 & 74     & 72 & 76        & 79  \\
Mistral-7B-Instruct-v0.1    & 62.77   & 73    & 70 & 62   & 24 & 87      & 43 & 69     & 36 & 79        & 84  \\
CodeLlama-34b-Instruct-hf   & 62.18   & 79    & 64 & 60   & 23 & 82      & 53 & 58     & 63 & 79        & 61  \\
c4ai-command-r-plus         & 60.77   & 79    & 60 & 63   & 8  & 72      & 72 & 66     & 59 & 78        & 50  \\
Meta-Llama-3-8B-Instruct    & 60.65   & 69    & 53 & 57   & 14 & 79      & 78 & 59     & 66 & 77        & 54  \\
starchat2-15b-v0.1          & 60.57   & 58    & 70 & 58   & 25 & 87      & 73 & 62     & 36 & 75        & 62  \\
gpt-4o                      & 58.58   & 60    & 63 & 68   & 2  & 71      & 77 & 47     & 61 & 75        & 62  \\
gpt-3.5-turbo               & 58.32   & 55    & 61 & 66   & 3  & 76      & 76 & 71     & 47 & 76        & 50  \\
Yi-1.5-34B-Chat             & 58.20   & 67    & 54 & 55   & 13 & 72      & 70 & 61     & 65 & 78        & 48  \\
Llama-2-7b-chat-hf          & 57.47   & 74    & 61 & 58   & 22 & 67      & 69 & 50     & 60 & 75        & 39  \\
zephyr-7b-alpha             & 56.96   & 57    & 62 & 47   & 23 & 85      & 29 & 44     & 73 & 76        & 75  \\
Meta-Llama-3-70B-Instruct   & 56.73   & 60    & 61 & 55   & 4  & 71      & 75 & 57     & 59 & 75        & 50  \\
gpt-4o-mini                 & 56.21   & 56    & 58 & 62   & 3  & 70      & 75 & 46     & 58 & 75        & 57  \\
Mistral-8x7B-Instruct-v0.1  & 52.80   & 60    & 53 & 48   & 23 & 71      & 46 & 43     & 59 & 73        & 52  \\
Mistral-7B-Instruct-v0.3    & 52.45   & 53    & 58 & 47   & 25 & 75      & 38 & 49     & 59 & 73        & 47  \\
Mistral-7B-Instruct-v0.2    & 50.18   & 13    & 58 & 54   & 14 & 72      & 61 & 46     & 64 & 71        & 49  \\
zephyr-7b-beta              & 47.85   & 28    & 53 & 48   & 26 & 71      & 7  & 38     & 73 & 75        & 60  
\end{tblr}
\caption{The humanlikness score for models in different experiments.}
\label{tab:models}
\end{table*}

\section{Result}

The overall humanlikeness scores revealed notable variations in how well LLMs emulated human language use across the 10 psycholinguistic experiments. Here, we performed a concise analysis to explore the data.

OpenAI’s models, including GPT-3.5-turbo, GPT-4o-mini, and GPT-4o, exhibited relatively stable performance across tasks, maintaining consistent humanlikeness scores. In contrast, the Llama family of models showed an overall increase in humanlikeness scores, with Meta-Llama-3.1-70B-Instruct achieving the highest performance among all Llama models. On the other hand, the Mistral family of models showed a slight decrease in humanlikeness, with Mistral-7B-Instruct-v0.3 scoring lower than its predecessors, indicating less alignment with human language use.


\subsection{Comparative Analysis}

Statistical comparisons between model families (three models selected per model family) revealed significant differences in performance. Notably, Llama models significantly outperformed Mistral models in humanlikeness (\textit{t} = 10.44, \textit{p} <.001) highlighting the substantial gap between these two families. Furthermore, Llama models also outperformed OpenAI models (\textit{t} = 3.13, \textit{p} =.002) although this difference was less pronounced compared to the Llama vs. Mistral comparison.

Within the Llama family, the transition from Meta-Llama-3-70B-Instruct to Meta-Llama-3.1-70B-Instruct showed a significant increase in humanlikeness (\textit{t} = -4.85, \textit{p} < .001), indicating improvements in model performance. In contrast, no significant differences were observed between GPT-3.5-turbo and GPT-4o (\textit{t} = -0.93, \textit{p} = 0.352), suggesting that OpenAI’s models performed consistently across experiments. Interestingly, within the Mistral family, Mistral-7B-Instruct-v0.3 showed a significant decrease in humanlikeness compared to Mistral-7B-Instruct-v0.1 (\textit{t} = 5.45, \textit{p} < .001).

These results underscore the varying abilities of different model families to approximate human language patterns, with Llama models demonstrating superior performance overall.

\subsection{Case analysis}
An in-depth analysis of individual experiments further highlights how LLMs’ performance varies in replicating human-like responses. Experiment 4, which tested word meaning priming, emerged as the most non-humanlike among the tasks, with substantial differences between human and LLMs’ responses (\textit{t} = -116.32, \textit{p} < .001). In this experiment, we assessed whether humans and models tend to access, when reading an ambiguous word such as post, the meaning previously used in the prime of an ambiguous word. Human participants exhibited a modest priming effect, with 20\% associating post with its job-related meaning after the word-meaning prime and 18\% after the synonym prime. In contrast, the Llama-3.1-70B model demonstrated a significantly higher priming effect, with 52\% responding to the word-meaning prime and 38\% to the synonym prime, revealing a stark divergence from human patterns. This case study emphasizes the challenges LLMs face in aligning their semantic associations with human interpretations, particularly when processing ambiguous or polysemous words. 
                                                 
\section{Discussion}
The results of this benchmark study highlight notable differences in how LLMs approximate human language use across various linguistic levels. The Llama family of models, particularly Meta-Llama-3.1-70B-Instruct, consistently outperformed both the OpenAI and Mistral models in terms of humanlikeness score. This finding suggests that recent advancements in the Llama models have led to more humanlike language behaviors, especially in terms of semantic and discourse processing.
The OpenAI models, including GPT-4o and GPT-3.5-turbo, showed relatively stable performance across tasks, with no significant differences between the models. This stability may reflect a plateau in the improvement of humanlikeness in these models, as compared to the more recent gains observed in the Llama family. On the other hand, the Mistral models demonstrated a decrease in humanlikeness scores, particularly in the transition to Mistral-7B-Instruct-v0.3. This suggests that certain training methods and data quality in Mistral may have reduced their alignment with human language patterns.
One of the key insights from this study is that models differ not only in their overall humanlikeness scores but also in how they handle specific linguistic phenomena. For instance, in Experiment 4 (word meaning priming), we observed a significant divergence in resposnes between humans and LLMs, with the latter showing a much larger priming effect. This over-priming suggests that while LLMs may excel in certain aspects of language generation, they often lack the subtle flexibility that humans display when processing ambiguous or context-dependent language.
A major strength of this study is its use of psycholinguistic experiments to evaluate LLMs, which goes beyond traditional NLP benchmarks that focus on task accuracy. By systematically probing various linguistic levels—sound, word, syntax, semantics, and discourse—this benchmark provides a more comprehensive understanding of how LLMs process and generate language.

\section{Conclusion}
In this paper, we introduced a novel benchmark for evaluating the humanlikeness of LLMs in language use based on psycholinguistic experiments. Our study evaluated 20 LLMs, including OpenAI’s GPT family, Meta’s Llama family, the Mistral family and others, across 10 experiments that spanned key linguistic aspects such as sound, word, syntax, semantics, and discourse. Using responses from over 2,000 human participants as a baseline, the results revealed significant differences in model performance, with Llama models consistently outperforming both OpenAI and Mistral models in terms of language use humanlikeness. These findings underscore the potential of psycholinguistic benchmarks to capture aspects of language that are often missed by traditional NLP evaluations.

This benchmark provides a framework for future research on LLMs, offering a more meaningful and comprehensive way to evaluate their performance in real-world language use. It also highlights areas where current LLMs diverge from human language patterns, particularly in tasks involving semantic priming and ambiguity resolution. By identifying these gaps, this study offers critical insights for the next generation of LLM development, paving the way for models that more closely mirror the intricacies of human communication.

\section{Limitation}
However, there are several limitations to this study. First, while the benchmark covers a wide range of linguistic tasks, it may not encompass the full complexity of human language use. Some linguistic phenomena, such as pragmatic reasoning, were not explored in this study. Second,  we did not manipulate models’ parameters, particularly the temperature or top k, to control the diversity of the generated responses. While using default parameters, particularly temperature, may seem limiting, this choice ensures that we evaluate models in their most typical and practical configurations. Default settings reflect how these models are commonly used in real-world applications, offering a fair and standardized comparison. Tuning parameters like temperature could introduce bias and variability across models, making it difficult to ensure consistent evaluation. By using default settings, we eliminate these concerns, allowing for a more reliable assessment of humanlikeness. Finally, while the study includes a large sample of human participants, the specific demographic characteristics (e.g., native English speakers from the UK and US) may not fully represent global language use patterns. Compared to previous benchmarks that focus on task-based performance, this study offers a more in-depth analysis of language models’ alignment with human linguistic behavior. Similar studies, such as Ettinger (2020), have used psycholinguistic principles to probe LLMs, but our study stands out by incorporating a broader range of linguistic levels and by using a large-scale dataset of human responses for direct comparison. The significant differences found between model families, such as the higher humanlikeness of Llama models, provide valuable insights for the ongoing development and fine-tuning of LLMs.

\label{sec:bibtex}


\bibliography{custom}

\appendix

\section{Appendix}
\label{sec:appendix}

This section introduces the ten psycholinguistic experiments used to evaluate the humanlikeness of LLMs across multiple linguistic levels. Each experiment was designed to test a specific linguistic phenomenon and compare the performance of LLMs to human participants.

\textit{\textbf{Sounds: sound-shape association}} People often associate specific sounds with certain shapes, a phenomenon known as sound symbolism. We tested whether LLMs, like humans, tend to link spiky-sounding words such as \textit{takete} or \textit{kiki} with spiky objects and round-sounding words like \textit{maluma} or \textit{bouba} with round objects.

\textbf{\textit{Sounds: sound-gender association}} People can often guess if an unfamiliar name is male or female based on its sound. In English, women’s names more frequently end in vowels compared to men’s names. In this task, we asked participants to complete a preamble containing either a consonant-ending name (e.g., \textit{Pelcrad} in 1a) or a vowel-ending novel name (e.g., \textit{Pelcra} in 1b).

1a. Consonant-ending name: \textit{Although Pelcrad was sick...}

1b. Vowel-ending name: \textit{Although Pelcra was sick...}

\textbf{\textit{Words: word length and predictivity}} Shorter words are suggested to make communication more efficient by carrying less information. If both humans and LLMs are sensitive to the relationship between word length and informativity, they should prefer shorter words over longer ones with nearly identical meanings when completing sentence preambles that predicted the meaning of the word (making it less informative; e.g., 2a), compared to neutral sentence preambles (e.g., 2b)

2a. Predictive context: \textit{Susan was very bad at algebra, so she hated... 1. math  2. mathematics}

2b. Neutral context: \textit{Susan introduced herself to me as someone who loved... 1. math    2. mathematics}

\textit{\textbf{Words: word meaning priming}} Many words have multiple meanings; for instance, \textit{post} can refer to mail or a job. People update an ambiguous word's meaning based on recent exposure. We tested whether humans and LLMs similarly demonstrate word meaning priming phennomenon: Participants associated post with its job-related meaning more frequently after reading sentences using that context rather than synonyms' contexts (3a vs.3b). 

3a. Word-meaning prime: \textit{The man accepted the post in the accountancy firm.}

3b. Synonym prime: \textit{The man accepted the job in the accountancy firm.}

\textbf{\textit{Syntax: structural priming}} In structural priming, people tend to repeat syntactic structures they've recently encountered. We had participants complete prime preambles designed for either PO (prepositional-object dative structure, e.g., \textit{The racing driver gave helpful mechanic wrench} to complete 4a) or DO (double-object dative structure, e.g., \textit{The racing driver gave torn overall his mechanic} to complete 4b). Participants then completed target preamble which could be continued as either DO/PO. If structural priming is demonstrated, participants replicate structure of the prime preamble.

4a. DO-inducing prime preamble: \textit{The racing driver showed the helpful mechanic ...}

4b. PO-inducing prime preamble: \textit{The racing driver showed the torn overall ...}

4c. Target preamble: \textit{The patient showed ...}

\textbf{\textit{Syntax: syntactic ambiguity resolution}} The way people parse words into syntactic structures has garnered significant attention in psycholinguistics. For instance, in VP/NP ambiguity (e.g., \textit{The ranger killed the poacher with the rifle}), people usually interpret the ambiguous prepositional phrase (PP, \textit{with the rifle}) as modifying the verb phrase (VP, \textit{killed the poacher}) rather than the noun phrase (NP, \textit{the poacher}). However, contextual information can modulate this resolution: People are more likely to interpret ambiguous PPs as modifying NPs when there are multiple possible referents (e.g., 5b) compared to when there is only a single referent (e.g., 5a). We examine how effectively LLMs use contextual information to resolve syntactic ambiguities and exhibit such modulation patterns.

5a. Single referent: \textit{There was a hunter and a poacher. The hunter killed the dangerous poacher with a rifle not long after sunset. Who had a rifle, the hunter or the poacher? }

5b. Multiple referents: \textit{There was a hunter and two poachers. The hunter killed the dangerous poacher with a rifle not long after sunset. Who had a rifle, the hunter or the poacher?}

\textbf{\textit{Meaning: implausible sentence interpretation}} Listeners often need to recover intended messages from noise-corrupted input. Errors in production or comprehension can make a plausible sentence implausible by omitting (e.g., \textit{to} omitted, 6a) or inserting words (e.g., \textit{to} inserted, 6b). People may interpret an implausible sentence nonliterally if they believe it is noise-corrupted. who found that people more frequently reinterpret implausible DO sentences than PO sentences due to the likelihood of omissions over insertions. We tested whether people and LLMs similarly assume that implausible sentences result from noise corruption, with omissions being more likely than insertions.

6a. Implausible DO: \textit{The mother gave the candle the daughter.}

6b. Implausible PO: \textit{The mother gave the daughter to the candle.}

6c. Question: \textit{Did the daughter receive something/someone?}

\textbf{\textit{Meaning: semantic illusions}} People often overlook obvious errors in sentences. For instance, when asked (7a), many fail to notice that the question should refer to \textit{Noah} instead of \textit{Moses}. Such semantic illusions suggest that processing sentence meanings involves partial matches in semantic memory. We tested whether LLMs and people alike produce semantic illusions and are more likely to catch a weak imposter (e.g., \textit{Adam}, less similar to \textit{Noah}, 7b) than a strong imposter (e.g. \textit{Morse}, more similar to \textit{Noah}, 7a). 

7a. Strong: \textit{During the Biblical flood, how many animals of each kind did Moses take on the ark? }

7b. Weak: \textit{During the Biblical flood, how many animals of each kind did Adam take on the ark?}

\textbf{\textit{Discourse: implicit causality}} Certain verbs prompt people to associate causality with either the subject or the object within a sentence. For instance, stimulus-experiencer verbs like scare typically lead people to attribute causality to the subject (e.g., completing 8a as \textit{Gary scared Anna because he was violent}), whereas experiencer-stimulus verbs like fear generally lead people to attribute causality to the object (e.g., completing 8b as \textit{Gary feared Anna because she was violent}). We assessed whether LLMs, like humans, show similar patterns of causal attribution based on verb type.

8a. Stimulus-experiencer verb: \textit{Gary scared Anna because...}

8b. Experiencer-stimulus verb: \textit{Gary feared Anna because...}

\textbf{\textit{Discourse: drawing inferences}} People make bridging inferences more frequently than elaborative inferences. Bridging inferences connect two pieces of information (after reading 9a, people infer that Sharon cut her foot) while elaborative inferences extrapolate from a single piece of information (people are less likely to make this inference after reading 9b). We examined how well an LLM aligns with human patterns of inference by comparing the bridging and elaborative conditions.

9a. Bridging: \textit{While swimming in the shallow water near the rocks, Sharon stepped on a piece of glass. She called desperately for help, but there was no one around to hear her.}

9b. Elaborative: \textit{While swimming in the shallow water near the rocks, Sharon stepped on a piece of glass. She had been looking for the watch that she misplaced while sitting on the rocks.}

Question: \textit{Did she cut her foot?}

\end{document}